\documentclass[12pt]{article}

\usepackage{amsmath}
\usepackage{amssymb}
\usepackage{float}
\usepackage[T1]{fontenc}
\usepackage{graphicx}
\usepackage{caption}
\usepackage[utf8]{inputenc}
\usepackage{mathtools}
\usepackage[normalem]{ulem}
\usepackage{wasysym}
\usepackage[hidelinks]{hyperref}

\title{Generating Clear Images From Images With Distortions Caused by Adverse Weather Using Generative Adversarial Networks}

\begin{document}
\maketitle

\begin{center}
\textbf{Nuriel Shalom Mor, Ph.D. \\ }\textit{Darca and Bnei Akiva Schools, Israel \\}
\end{center}

\begin{center}
\textit{With the participation of students of the artificial intelligence program 
Hadash-Darca, Bat Yam, and Harel-Bnei Akiva, Holon, Israel\\ }
\end{center}

\textbf{Abstract.} We presented a method for improving computer vision tasks on images affected by adverse weather conditions, including distortions caused by adherent raindrops. Overcoming the challenge of applying computer vision to images affected by adverse weather conditions is essential for autonomous vehicles utilizing RGB cameras. For this purpose, we trained an appropriate generative adversarial network and showed that it was effective at removing the effect of the distortions, in the context of image reconstruction and computer vision tasks. We showed that object recognition, a vital task for autonomous driving vehicles, is completely impaired by the distortions and occlusions caused by adherent raindrops and that performance can be restored by our de-raining model. The approach described in this paper could be applied to all adverse weather conditions.  

\section{Introduction}

Autonomous vehicles can make use of various sensors (Mușat, Fursa, Newman, Cuzzolin, $\&$ Bradley, 2021). Currently, there is a heavy reliance on vision-based sensors such as RGB cameras (Mușat et al., 2021). RGB cameras are affected by adverse weather (Porav, Bruls, $\&$ Newman, 2019; Zheng, Wu,  Han , $\&$ Shi, 2020).  Therefore, the performance of autonomous vehicles is severely affected by adverse weather, ultimately, compromising the safety of the traffic participants (Zhang, Carballo,  Yang,  $\&$ Takeda, 2021).

\begin{flushleft}
Because of the recent rise of the autonomous driving industry, much effort has been devoted to tackling this issue (Zang et al., 2019). Hardware solutions are being developed but are limited in their ability to mitigate the problem when relying mainly on RGB cameras. When rain, fog, snow, smog, or sand interact with cameras’ lenses, the task of computer vision becomes challenging. This problem prevents autonomous vehicles which rely on RGB cameras to ensure a safe journey (Fursa, et al.,2021; Hassaballah, Kenk, Muhammad, $\&$ Minaee, 2020; Yoneda, Suganuma, Yanase, $\&$ Aldibaja, 2019).

\end{flushleft}

\begin{flushleft}
Adverse weather conditions can have an impact on the quality of an image in two ways.  Firstly, rain, fog, smog, snow, or sandstorm might hinder visibility or occlude a scene but do not distort the image. Secondly, water droplets, sticking to lenses, distort the image, acting as secondary lenses with various degrees of blurring and distortions (Porav et al., 2019; Zang, Ding,  Smith, Tyler, Rakotoarivelo, $\&$ Kaafar, 2019). 
 
\end{flushleft}

\begin{flushleft}

Therefore, in this study, we focused on solving the problem of images with distortions caused by heavy rain, adherent rain droplets, and adherent streaks. The approach described in this paper can easily be applied to the effect of fog, smog, snow, and sandstorm. Heavy rain, adherent rain droplets, and adherent streaks present a double challenge, not only hindering and occluding the scene but also distorting the image. Other bad weather conditions such as snow, smog, fog, and sandstorm only limit the field of view and do not distort the image. 
\end{flushleft}

\begin{flushleft}
Image distortions caused by adherent rain droplets impede all computer vision tasks, including object recognition, localization, and semantic segmentation which are essential for autonomous vehicles (Porav et al., 2019). We built a system in which the input was images with significant distortions caused by adherent rain droplets, and the output was cleared images that improve the performance of computer vision tasks performed on the images. 
\end{flushleft}

\begin{flushleft}
Our approach was to obtain a dataset of rain distorted - clear image pairs and train a Generative Adversarial Network (GAN) to predict clear images from rain-distorted images. The idea was that the model would learn how to predict clear images from rain-distorted images. The distortions caused by adherent rain droplets are not random noise. The GAN could learn the relationship between distorted images and clear images. 
\end{flushleft}

\begin{flushleft}

This approach is a practical innovative alternative to the traditional supervised deep learning approach of acquiring huge amounts of rainy images, annotating, and labeling them for object recognition and semantic segmentation. The traditional supervised learning approach in this case is time-consuming, expensive, and almost impossible (Porav et al., 2019). The purpose of this study was to construct a de-raining model that produces state-of-the-art results which could be further developed for autonomous vehicles utilizing cameras. 

\end{flushleft}

\subsection{Generative Adversarial Networks (GANs)}

\begin{flushleft}
GANs are the state-of-the-art framework for training generative models. No inference is required during learning avoiding the difficulty of approximating intractable probabilistic computations (Creswell et al., 2018; Mirza $\&$ Osindero, 2014). 

\end{flushleft}

\begin{flushleft}
GANs consist of two adversarial neural networks: 
1. a generative network that captures the data distribution.
2. a discriminative model that estimates the probability, a sample is from the training data rather than the generator (Isola, Zhou, $\&$ Efros, 2017). 

\end{flushleft}

\begin{flushleft}
The generator and the discriminator can be deep neural networks and more specifically convolutional neural networks (CNNs; Isola et al., 2017). The domain of computer vision using CNNs is one of the areas that progressed dramatically during the deep learning era with many different types of applications (Mor et al., 2021). The generator and the discriminator are trained simultaneously to adjust both parameters following a two-player min-max game.  
\end{flushleft}

\begin{flushleft}

The generator produces images and the task of the discriminator is to distinguish between real images and images that were generated by the generator. The aim of the discriminator is to assign the correct label to the real example and the example generated by the generator (real vs. generated). The aim of the generator is to “outsmart" the discriminator, so it would label generated images as real (Creswell et al., 2018; Mirza $\&$ Osindero, 2014). 

\end{flushleft}

\begin{flushleft}
\textit{\textbf{pix2pix network.}} \textit{pix2pix} is a cGANs able to perform high-resolution image-to-image translation. This network learns the mapping from an input image to an output image and a loss function to train this mapping (Isola et al., 2017). Conditional GAN is a type of GAN in which we can direct the data generation process based on extra information such as class labels (Atienza, 2019; Peters, $\&$ Brenner, 2020). 

\end{flushleft}

\begin{flushleft}
The purpose of this research was to examine whether \textit{pix2pix} could generate clear images from rain-distorted images given appropriate training data. Clear images that would improve the performance of computer vision tasks performed on the images.  
\end{flushleft}

\section{Method}

\subsection{Dataset}

\begin{flushleft}
We selected an appropriate data set for the purpose of this study: to train a GAN to predict clear images from rain-distorted images (Porav et al., 2019). The data set contained about 50000 pairs of images, rain distorted - clear image pairs.  They (Porav et al., 2019) created a real-world stereo dataset where one lens is affected by real water droplets and the other is kept dry. We used 40000 image pairs to form a training dataset, 500 image pairs to form a validation data set, and 500 image pairs to form a test data set. An example of the image pairs in the data set is shown in figure 1. 
\end{flushleft}

\begin{flushleft}
In this study, input images refer to the images with distortions caused by raindrops. Predicted images are the generated clear images. Ground truth refers to the original clear images that are aligned with the input. 
\end{flushleft}

The dataset is available at \\ \url{https://drive.google.com/uc?id=1j_5ho1atUIXrufzCwwmPhpkKn440AN0d&export=download} 

\begin{figure}[H]
\raggedright
\includegraphics[width=11.38cm,height=7.06cm]{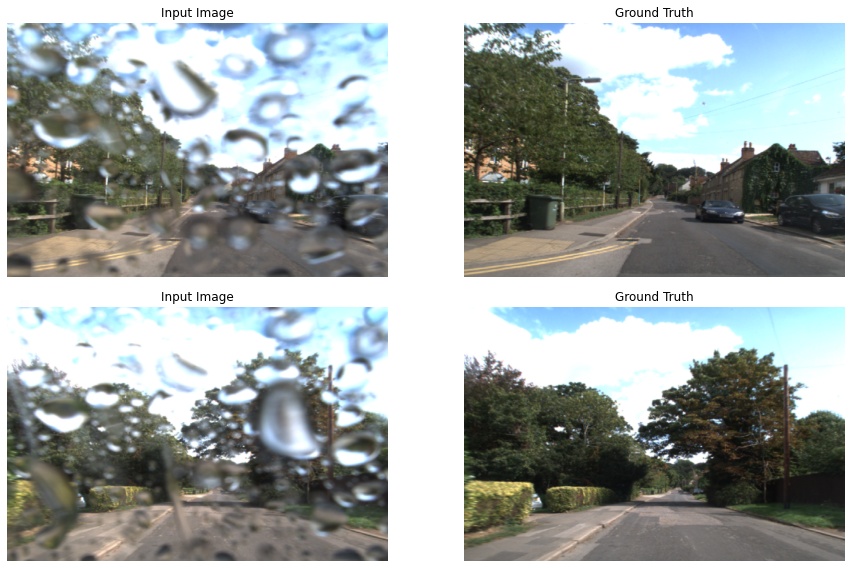}
\caption{Examples of aligned pairs from the dataset. On the left, input images that are affected by real raindrops (input image). On the right, original aligned clear images (ground truth).   
}
\label{fig:}
\end{figure}

\subsection{Training Procedure}

\begin{flushleft}
Our training procedure was inspired by \textit{pix2pix} with the required technical modifications and adjustments (Isola et al., 2017). Our generator and discriminator architectures are adapted from Isola et al., 2017. Both the generator and discriminator use modules of the form convolution-BatchNorm-ReLu. 
\end{flushleft}

\subsection{Performance Measurement}

\begin{flushleft}
The task of evaluating GANs’ performance is complicated. There is no
objective function used when training GAN generator models, meaning
models must be evaluated using the quality of the generated synthetic
images (Alqahtani et al., 2019; DeVries et al., 2019). Manual inspection of
generated images is a good starting point when getting started. Quantitative measures, such as the inception score and the Frechet inception distance can evaluate the quality of the generated images (Alqahtani et al., 2019; DeVries et al., 2019).

\end{flushleft}

\begin{flushleft}
We needed a measurement that would evaluate the performance of computer vision tasks on the original distorted images, the generated clear images, and the original clear images, in order to compare and assess any improvement in the generated clear images compared to the original distorted images. Therefore, we developed (Mor et al., 2021) a measure for our specific purpose. For each trio, derived from the test set, we calculated the number of cars that could be detected in the original distorted image, in the clear generated image, and in the original clear image using OpenCV, a standard computer vision object recognition software. 
\end{flushleft}

\begin{flushleft}
We calculated two terms. Term 1 is the number of cars that could be detected in the original distorted images divided by the number of cars that could be detected in the original clear images. Term 2 is the number of cars that could be detected in the clear generated images divided by the number of cars that could be detected in the original clear images. The expectation is that these terms would be between 0 to 1. A value closer to 1 indicates better performance on visual object recognition tasks. In addition, a comparison between term 1 and term 2 could help us evaluate if there is an improvement in the performance of computer vision tasks performed on the generated images. As shown in the following formulas. \textit{m} is the number of examples. $n_c$  is the number of cars detected in the original clear images. $n_d$ is the number of cars detected in the original rain-distorted images. $n_p$ is the number of cars detected in the predicted (generated) clear images. 
 
\end{flushleft}

\begin{equation}
\label{eq:nolabel}
\frac{1}{m}\sum \frac{n_d}{n_c}
\end{equation}

\begin{equation}
\label{eq:nolabel}
\frac{1}{m}\sum \frac{n_p}{n_c}
\end{equation}

\section{Results And Experiments}

\begin{flushleft}
The input images contain distortions caused by raindrops. The ground truth is the alligned clear images. We trained the GANs for 34 epochs with a batch size of one. After 34 epochs the generator started to overfit the input images. Figures 2, 3, 4. 5. 6, 7, and 8 show an example of predicted images from the test set. The test set is a completely new recording that the GANs have never seen before the test. 

\end{flushleft}

\begin{flushleft}
Our purpose was to evaluate whether \textit{pix2pix} could generate clear images from images with distortions caused by raindrops, which improve the performance of computer vision tasks performed on the images. As mentioned, in order to evaluate the improvement in the performance of computer vision tasks on the clear-predicted images compared to the distorted input images, we developed a measure for this specific goal.  For each pair on the test set, we calculated the number of cars that could be detected in the predicted image divided by the number of cars that could be detected in the ground truth. In addition, we calculated the number of cars that could be detected in the input image divided by the number of cars present in the ground truth.  Summed each term for all examples and divided it by the number of examples. The closer the score is to 1 indicates better performance of computer vision tasks.  

\end{flushleft}

\begin{flushleft}
We found that for the predicted images the score was 0.94 and for the input images the score was 0.12. These results suggest that raindrops distorted the images to the extent that visual object recognition could not be applied. In addition, these results suggest that the generated images are almost identical to the original clear images in terms of computer vision tasks. Our de-raining model produces extraordinary results on our test set, restoring the performance of object visual recognition tasks on images affected by adherent raindrops.

\end{flushleft}

\begin{figure}[H]
\raggedright
\includegraphics[width=11.38cm,height=7.06cm]{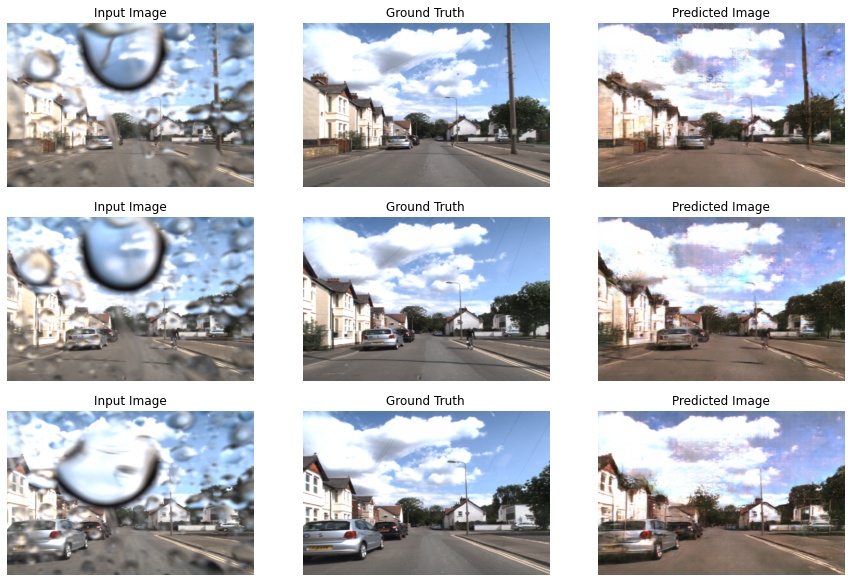}
\caption{On the left, input images that are affected by raindrops. In the middle, the original aligned clear images (ground truth).  On the right, the generated images (predicted image)  
}
\label{fig:}
\end{figure}

\begin{figure}[H]
\raggedright
\includegraphics[width=11.38cm,height=7.06cm]{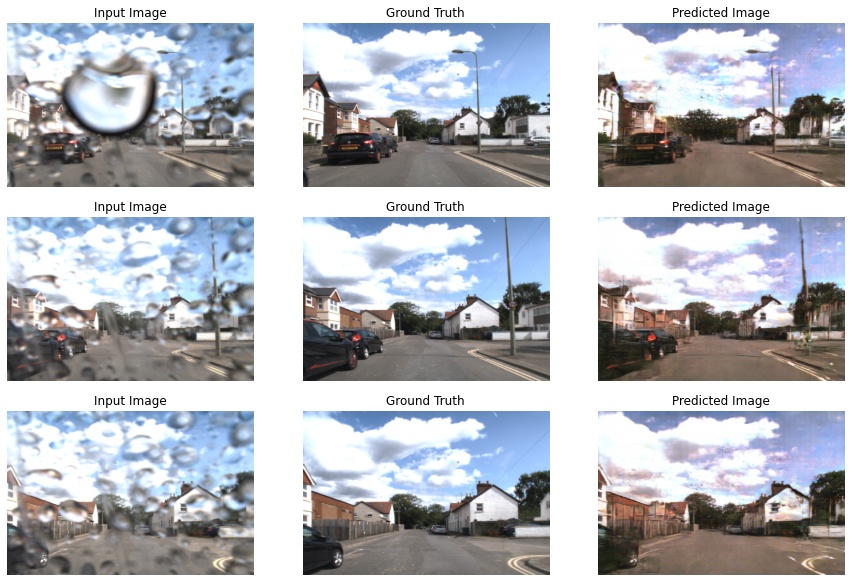}
\caption{On the left, input images that are affected by raindrops. In the middle, the original aligned clear images (ground truth).  On the right, the generated images (predicted image)  
}
\label{fig:}
\end{figure}

\begin{figure}[H]
\raggedright
\includegraphics[width=11.38cm,height=7.06cm]{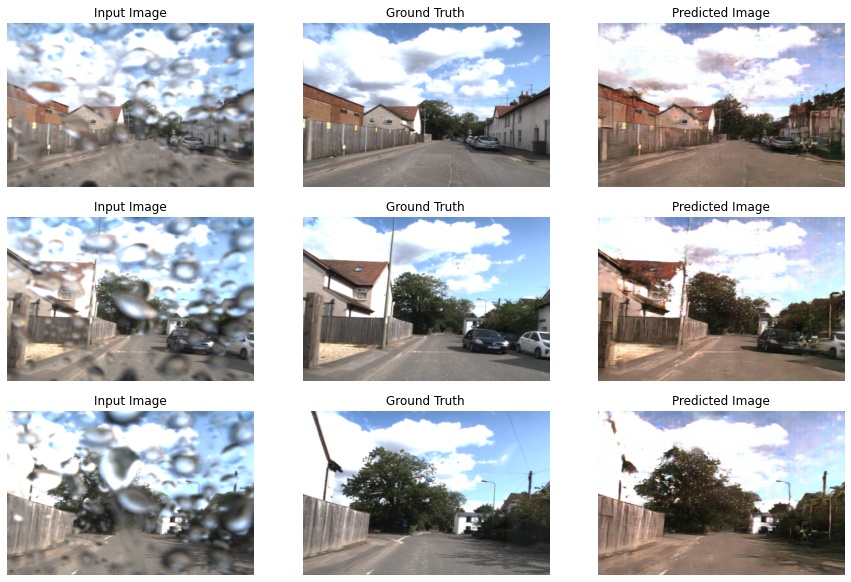}
\caption{On the left, input images that are affected by raindrops. In the middle, the original aligned clear images (ground truth).  On the right, the generated images (predicted image)  
}
\label{fig:}
\end{figure}

\begin{figure}[H]
\raggedright
\includegraphics[width=11.38cm,height=7.06cm]{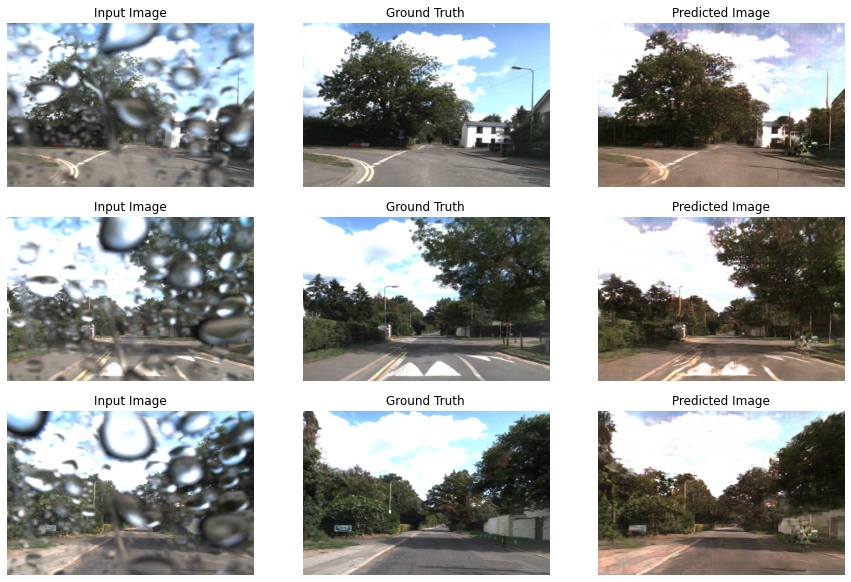}
\caption{On the left, input images that are affected by raindrops. In the middle, the original aligned clear images (ground truth).  On the right, the generated images (predicted image)  
}
\label{fig:}
\end{figure}

\begin{figure}[H]
\raggedright
\includegraphics[width=11.38cm,height=7.06cm]{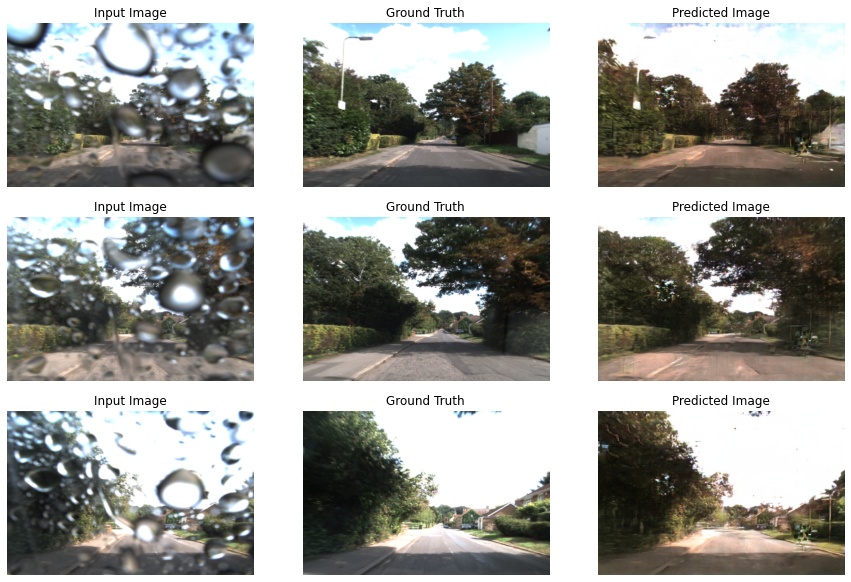}
\caption{On the left, input images that are affected by raindrops. In the middle, the original aligned clear images (ground truth).  On the right, the generated images (predicted image)  
}
\label{fig:}
\end{figure}

\begin{figure}[H]
\raggedright
\includegraphics[width=11.38cm,height=7.06cm]{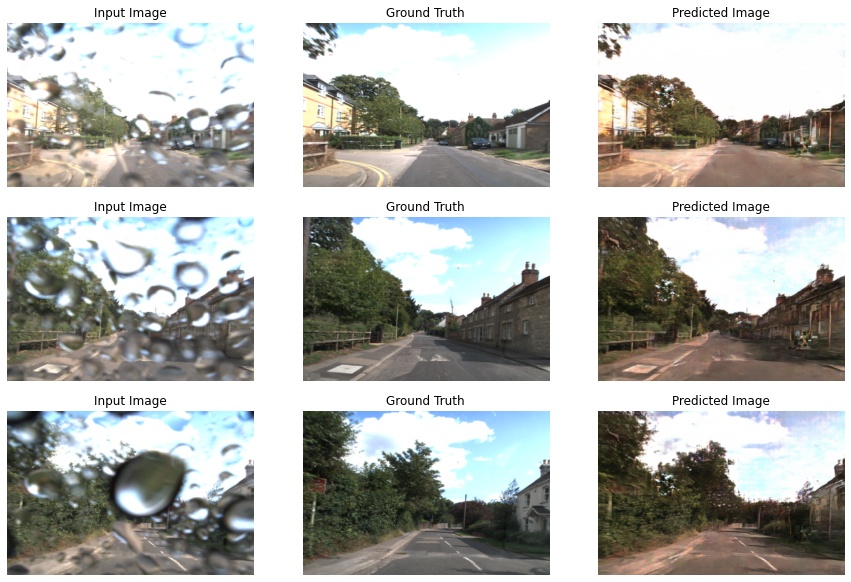}
\caption{On the left, input images that are affected by raindrops. In the middle, the original aligned clear images (ground truth).  On the right, the generated images (predicted image)  
}
\label{fig:}
\end{figure}

\begin{figure}[H]
\raggedright
\includegraphics[width=11.38cm,height=7.06cm]{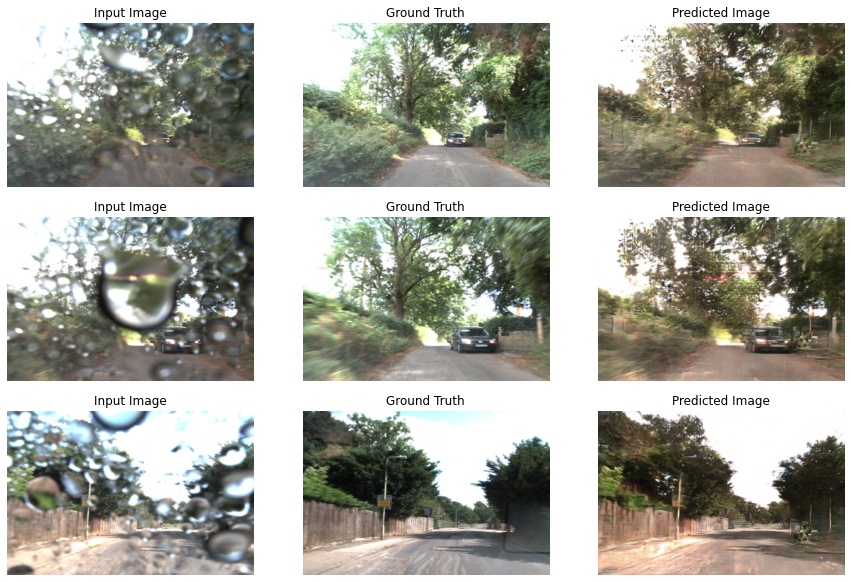}
\caption{On the left, input images that are affected by raindrops. In the middle, the original aligned clear images (ground truth).  On the right, the generated images (predicted image)  
}
\label{fig:}
\end{figure}

\subsection{Conclusion}

\begin{flushleft}
We presented a de-raining model that restores the performance of computer vision tasks on images affected by adherent raindrops. Our results show that object recognition, a vital task for autonomous driving vehicles, is completely impaired by the distortions and occlusions caused by adherent raindrops and that the performance can be restored by our de-raining model. 
\end{flushleft}

\begin{flushleft}
The approach described in this paper could be applied to all adverse weather conditions including, fog, smog, snow, and sandstorm. All of these adverse weather conditions hinder visibility or occlude a scene but do not distort the image. Adherent raindrops sticking to lenses, not only occlude a scene but also distort the image, acting as secondary lenses with various degrees of blurring and distortions. 
\end{flushleft}

\begin{flushleft}
Our model is available at \\ \url{https://drive.google.com/drive/folders/18snuPC2TQ_SsUrTOD5xaHR0r1B3jbgz_?usp=share_link}
\end{flushleft}

\section{Authors’ Note}
Dedicated to the memorial of my father, Moshe Mor, peace be upon him.
\begin{flushleft}
Please send correspondence to Nuriel S. Mor, Ph.D. Software engineering program with an emphasis on AI, Darca and Bnei Akiva schools, Israel. Email: \href{mailto:nuriel.mor@gmail.com}{\uline{nuriel.mor@gmail.com}}
\end{flushleft}
The idea for such a system, as described in this article, was first proposed and developed by Dr. Mor in October 2022

\section{References}

\begin{flushleft}
Alqahtani, H., Kavakli-Thorne, M., Kumar, G., $\&$ SBSSTC, F. (2019, December). An analysis of evaluation metrics of gans. In International Conference on Information Technology and Applications (ICITA).
\end{flushleft}

\begin{flushleft}
Atienza, R. (2019). A conditional generative adversarial network for rendering point clouds. In Proceedings of the IEEE/CVF Conference on Computer Vision and Pattern Recognition Workshops (pp. 10-17).
\end{flushleft}

\begin{flushleft}
Creswell, A., White, T., Dumoulin, V., Arulkumaran, K., Sengupta, B., $\&$ Bharath, A. A. (2018). Generative adversarial networks: An overview. \textit{IEEE Signal Processing Magazine}, \textit{35}(1), 53-65.
\end{flushleft}

\begin{flushleft}
DeVries, T., Romero, A., Pineda, L., Taylor, G. W., $\&$ Drozdzal, M. (2019). On the evaluation of conditional gans. \textit{arXiv preprint arXiv:1907.08175}.
\end{flushleft}

\begin{flushleft}
Fursa, I., Fandi, E., Musat, V., Culley, J., Gil, E., Teeti, I., ... $\&$ Bradley, A. (2021). Worsening perception: Real-time degradation of autonomous vehicle perception performance for simulation of adverse weather conditions. \textit{arXiv preprint arXiv:2103.02760}
\end{flushleft}

\begin{flushleft}
Hassaballah, M., Kenk, M. A., Muhammad, K., $\&$ Minaee, S. (2020). Vehicle detection and tracking in adverse weather using a deep learning framework. \textit{IEEE transactions on intelligent transportation systems, 22(7), 4230-4242}
\end{flushleft}

\begin{flushleft}
Isola, P., Zhu, J. Y., Zhou, T., $\&$ Efros, A. A. (2017). Image-to-image translation with conditional adversarial networks. In \textit{Proceedings of the IEEE conference on computer vision and pattern recognition} (pp. 1125-1134).
\end{flushleft}

\begin{flushleft}
Mirza, M., $\&$ Osindero, S. (2014). Conditional generative adversarial nets. \textit{arXiv preprint arXiv:1411.1784}
\end{flushleft}

\begin{flushleft}
Mor, N. S. (2021). Generating Photo-realistic Images from LiDAR Point Clouds with Generative Adversarial Networks. \textit{arXiv preprint arXiv:2112.11245.}
\end{flushleft}

\begin{flushleft}
Mor, N. S., $\&$ Dardeck, K. L. (2020). Applying Deep Learning to Specific Learning Disorder Screening. \textit{arXiv preprint arXiv:2008.13525}.
\end{flushleft}

\begin{flushleft}
Mușat, V., Fursa, I., Newman, P., Cuzzolin, F., $\&$ Bradley, A. (2021). Multi-weather city: Adverse weather stacking for autonomous driving. \textit {In Proceedings of the IEEE/CVF International Conference on Computer Vision (pp. 2906-2915).}
\end{flushleft}

\begin{flushleft}
Peters, T., $\&$ Brenner, C. (2020). Conditional adversarial networks for multimodal photo-realistic point cloud rendering. \textit{PFG–Journal of Photogrammetry, Remote Sensing and Geoinformation Science}, \textit{88}(3), 257-269.

\end{flushleft}

\begin{flushleft}
Porav, H., Bruls, T., $\&$ Newman, P. (2019, May). I can see clearly now: Image restoration via de-raining. \textit{In 2019 International Conference on Robotics and Automation (ICRA) (pp. 7087-7093). IEEE.}

\end{flushleft}

\begin{flushleft}
Yoneda, K., Suganuma, N., Yanase, R., $\&$ Aldibaja, M. (2019). Automated driving recognition technologies for adverse weather conditions. \textit{IATSS research, 43(4), 253-262.}

\end{flushleft}

\begin{flushleft}
Zang, S., Ding, M., Smith, D., Tyler, P., Rakotoarivelo, T., $\&$ Kaafar, M. A. (2019). The impact of adverse weather conditions on autonomous vehicles: How rain, snow, fog, and hail affect the performance of a self-driving car. \textit {IEEE vehicular technology magazine, 14(2), 103-111.}

\end{flushleft}

\begin{flushleft}

Zhang, Y., Carballo, A., Yang, H., $\&$ Takeda, K. (2021). Autonomous Driving in Adverse Weather Conditions: A Survey. \textit{arXiv preprint arXiv:2112.08936.}
\end{flushleft}

\begin{flushleft}
Zheng, Z., Wu, Y., Han, X., $\&$ Shi, J. (2020, August). Forkgan: Seeing into the rainy night. \textit{In European conference on computer vision (pp. 155-170). Springer, Cham.}
\end{flushleft}

\end{document}